\documentclass[10pt,journal,compsoc]{IEEEtran}
\usepackage{amsmath,amsfonts}
\usepackage{algorithmic}
\usepackage{algorithm}

\usepackage{array}
\usepackage{mathtools} % 在导言区添加
\usepackage[caption=false,font=normalsize,labelfont=sf,textfont=sf]{subfig}
\usepackage{textcomp}
\usepackage{stfloats}
\usepackage{url}
\usepackage{verbatim}
\usepackage{graphicx}
\usepackage{cite}
\usepackage{float} 
\usepackage{caption} 
\usepackage[numbers,sort&compress]{natbib}
\usepackage{booktabs} % 用于生成符合学术标准的三线表（\toprule, \midrule, \bottomrule）
\usepackage{multirow} % 用于合并行
\begin{document}

\title{A Generalized Deep Non-negative Matrix Factorization Approach for SAR Automatic Target Recognition}

\author{Yunhong Zhang, Changjie Cao*, Member, IEEE, Zhongli Zhou, Bingli Liu, Zongjie Cao, Member, IEEE, Zongyong Cui, Member, IEEE, Ying Yang
	 \IEEEmembership{}
	% <-this % stops a space
	\thanks{This work was supported by the National Science and Technology Major Projects of China(2025ZD1008203), the National Natural Science Foundation of China(42572389), the Sichuan Natural Science Foundation(2026NSFSC0238)}}

\markboth{IEEE TRANSACTIONS ON PATTERN ANALYSIS AND MACHINE INTELLIGENCE}%
{Shell \MakeLowercase{\textit{et al.}}: A Sample Article Using IEEEtran.cls for IEEE Journals}

\IEEEpubid{}
\IEEEtitleabstractindextext{%
\begin{abstract}
	The deep nonnegative matrix factorization (DNMF) technique is proposed to address the low interpretability of deep learning-based methods in extracting multilayer features from synthetic aperture radar (SAR) target samples. However, existing DNMF methods employ a layer-by-layer decomposition strategy, which is prone to causing error accumulation and local optimum, thereby hindering a consistent improvement in recognition accuracy as the number of layer increases. In this paper, a robust multilayer feature extraction method, termed generalized deep non-negative matrix factorization (G-DNMF), is proposed to address the above challenges in SAR automatic target recognition (ATR). The G-DNMF aims global optimality and derives the update rules for each parameter using lagrangian multiplier method. The new update formula indicates that both the DNMF method based on the encoding matrix and the mixing matrix are special cases of the proposed method, theoretically demonstrating the universality of proposed method. In general, the proposed method discards the layer-by-layer decomposition strategy, thereby effectively mitigating the risk of local optima and eliminating error accumulation, leading to a significant improvement in DNMF's multi-layer feature extraction capability. The experimental results, by presenting the feature images extracted from each layer by G-DNMF and the reconstructed original images, verified the proposed method's pure additive understanding of multi-layer features and demonstrated its interpretability. The experimental results based on MSTAR and OpenSARship datasets show that G-DNMF outperforms existing DNMF algorithms and their derivatives in terms of stability and recognition performance.
	
\end{abstract}

}

\maketitle

\section{Introduction}
\IEEEPARstart{S}{ynthetic} aperture radar(SAR) can penetrate cloud cover and operate continuously regardless of time or weather conditions, delivering essential data for earth observation \cite{ref1,ref2,ref3}. Among the many applications of SAR, the automatic target recognition (ATR) technology is the critical mean of SAR image interpretation \cite{ref4,ref5,ref6}. In general, the performance of SAR ATR methods is heavily contingent upon the selection of data representation strategy. The conventional approach to acquiring SAR target data representation is through feature extraction, which involves the design of elaborate algorithms for implementing transformations on SAR target data in support of ATR tasks \cite{ref7,ref8,ref9}. Among them, deep learning algorithms have become a important area of research by progressively constructing feature representations with discriminability and generalization capabilities \cite{ref10,ref11,ref12}, leveraging multi-layer representations of target samples \cite{ref13,ref14,ref15}.

However, the security implications of applying deep learning methods, such as convolutional neural networks (CNNs), to SAR ATR have become increasingly concerning due to lack transparency in internal logic processes \cite{ref16,ref17,ref18,ref19}. Specifically, SAR imagery pixel values are inherently positive, whereas CNNs-extracted features can take on both positive and negative values. These features clearly do not align with the human understanding of SAR imagery content, which is based on the hierarchical additive understanding of nonlinear transformation. Therefore, conventional CNN algorithms often lack transparency in explaining their feature extraction mechanisms, making it difficult to judge the root causes of misrecognition results \cite{ref20,ref21}.

To address this challenge, recent studies have proposed an approach known as deep non-negative matrix factorization (DNMF), which integrates interpretability with multi-layer feature extraction\cite{ref22,ref23,ref24,ref25,ref26}. DNMF factorizes original SAR data features layer by layer, constructing hierarchical feature representations. The layered factorization strategy not only uncovers the inherent feature structure within the data but also offers a transparent interpretation of the ATR model\cite{ref27}, allowing humans to intuitively understand the model's decision-making basis \cite{ref28}. Moreover, this architecture achieves interpretability and local feature representation by performing multi-layer factorization of the basis image matrix to extract underlying basis images \cite{ref29}. These characteristics make DNMF highly promising in scenarios where high interpretability and recognition accuracy are required, especially in SAR ATR tasks \cite{ref30}.

The current study focuses on leveraging DNMF to derive more discriminative and robust feature representations. Applying sparsity constraints to DNMF represents a straightforward and effective strategy. Motivated by the aforementioned viewpoints, Fang et .al utilized the $l_1$-norm as a sparsity penalty term for the coding matrix, thereby enhancing the sparsity of the DNMF \cite{ref42}. Analogously, Feng et al. introduced the $l_{1/2}$ regularization to effectively extract spectral and spatial features from hyperspectral images \cite{ref44}. To achieve stronger sparsity constraints, Huang et al. proposed a robust deep nonnegative matrix factorization method based on the $l_{2,1}$-norm ($l_{2,1}$-RDNMF) for hyperspectral unmixing \cite{ref34}. This method enhanced robustness to noise by incorporating the $l_{2,1}$-norm during both the pretraining and fine-tuning stages. However, the $l_{2,1}$-norm constraints also increases the computation complexity of the recognition model. To reduce the computational cost of $l_{2,1}$-RDNMF, Cao et .al proposed a deep incremental nonnegative matrix factorization method with $l_{2,1}$-norm constraint ($l_{2,1}$-DINMF) for SAR imagery recognition \cite{ref29}. $l_{2,1}$-DINMF significantly reduces the time and memory consumption in multi-layer feature extraction through incremental update rules. Guan et al. proposed Truncated CauchyNMF, using a truncated Cauchy loss to filter severe outliers and improve robustness \cite{ref48}. Gillis et al. proposed a distributionally robust multi-objective NMF method to optimally balance accuracy and robustness under data uncertainty \cite{ref49}. Additionally, Zhang et al. proposed a multiple non-linear activation-based deep non-negative matrix factorization approach for SAR automatic target recognition, which enhances feature discriminability by introducing non-linear activations and modifying the mixed matrix rather than the coding matrix \cite{ref50}.

In addition to imposing sparsity constraints, some DNMF approaches also attempt to draw on the principles or modules from deep learning to obtain better feature representations. For instance, Michel et al. iteratively applied semi-non-negative matrix factorization with layer-wise activation functions to the coding matrix, interpreting each layer's features as clustering to capture complex data structures \cite{ref22}. Ren et al. incorporated non-negativity constraints into both the encoder and decoder to better capture the intrinsic geometric structure of node pairs \cite{ref23}. Subsequently, Parekh et al. proposed a multi-view graph regularized deep autoencoder-based nonnegative matrix factorization (MGANMF) framework for multi-view clustering tasks \cite{ref28}. This framework integrates the hierarchical feature extraction ability and the interpretability, preserves the geometric structure of the data via graph regularization, and balances the contributions of different views through self-updating weighting mechanisms. Similarly, Tian et al. proposed a efficient deep matrix factorization architecture for the classification. The architecture integrated robust deep matrix factorization (RDMF) and dual-angle feature decomposition (DAFD), enabling more discriminative feature representations through sparse representation \cite{ref46}. Zhai et al. proposed a quadratic matrix factorization framework for manifold learning, which extracts intricate nonlinear features by explicitly modeling curved data manifolds through quadratic constraints \cite{ref47}.

Following extensive prior research, DNMF has demonstrated significant advantages in feature representation and interpretability \cite{ref31}. However, the aforementioned DNMF-based ATR models exhibit a notable performance degradation when the number of matrix factorization layers exceeds three. This phenomenon is primarily attributed to the inherent mathematical properties of NMF and the limitations of layer-wise factorization. Specifically, Non-negative Matrix Factorization is inherently a non-convex optimization problem, characterized by the existence of multiple local optima within its solution space \cite{ref26}. While the layer-wise factorization process enables progressive extraction of increasingly deeper features from the original data, the factorization results at each layer may become trapped near local optima, thereby restricting the optimization space for subsequent layers. The cumulative effect of these local optima is prone to causing more error accumulation, making it challenging for DNMF to proceed with deeper factorization \cite{ref32,ref33,ref34}. In summary, if the first layer of DNMF fails to fully capture the underlying features of the original data, subsequent layers must carry out further factorization based on these incomplete representations, leading to degraded feature expression and ultimately impairing the overall performance of the ATR model. To address the issues associated with layer-wise factorization, most DNMF algorithms employ a fine-tuning strategy to improve the performance of ATR models \cite{ref35,ref36,ref37}. It is noteworthy that fine-tuning may undermine the meaningful features acquired during the factorization process and introduce increased computational complexity \cite{ref38,ref39,ref40}. Moreover, this strategy is designed to mitigate the cumulative error inherent in multi-layer factorization, rather than reducing the risk of converging to local optima during the factorization process.

Motivated by the limitations of layer-wise factorization in DNMF, this paper proposes a novel feature extraction strategy that avoids layer-wise factorization, called generalized deep non-negative matrix factorization (G-DNMF). The following are the innovations and advantages of the proposed method:

\begin{itemize}
	\item{Unlike the conventional layer-wise factorization strategy, G-DNMF abandons the local optimization paradigm inherent in DNMF and instead adopts a global perspective to derive a novel iterative update rule.}
	
	\item{The novel iterative update rule theoretically proves that both the encoding matrix-based DNMF and the mixing matrix-based Deep NMF are special forms of the G-DNMF method. This brand-new universal understanding offers substantial potential for addressing the overarching challenges of DNMF and its derivative methods in future research and applications.}

	\item{Each parameter update in G-DNMF is jointly determined by all other parameters. 	This collaborative optimization mechanism among parameters strengthens parameter interaction and overcomes the limitation of unidirectional information flow inherent in conventional DNMF, where feature representation information is restricted to transmission from earlier layers to subsequent layers.}

	\item{During the layer-wise factorization process, the G-DNMF simultaneously avoids local optima and error accumulation, thereby providing a fundamental support for DNMF and its derivative methods to truly achieve deep feature representation.}

\end{itemize}

The rest of this paper is organized as follows. In section II, this paper surveys the related work on traditional DNMF. In section III, this paper elaborates on the proposed method, including the underlying ideas and the mathematical derivation of iterative formula. In section IV, this paper has carried out the necessary experimental design and verification and offered a comparative analysis of its performance against DNMF and its derivative algorithms. In section V, this paper discusses the conclusions. Finally, the appendix provides a detailed proof of the algorithm convergence of the proposed method.

\section{RELATED WORK}
\subsection{Encoding maxtrix-based DNMF}

In Non-negative Matrix Factorization (NMF), given a non-negative SAR image data matrix $\mathrm{V}\in \mathbb{R} ^{\mathrm{m}\times \mathrm{n}}$, the objective is to factorize it into a mixing matrix $\mathrm{W}\in \mathbb{R} ^{\mathrm{m}\times \mathrm{k}}$, and a coding matrix $\mathrm{H}\in \mathbb{R} ^{\mathrm{k}\times \mathrm{n}}$, such that $\mathrm{V}\approx \mathrm{WH}$. In most research related to Deep NMF, the coding matrix $H$ is further factorized. The coding matrix of each layer serves as the input for the next layer, thereby gradually extracting more advanced feature representations. The structure of Deep NMF based on encoding matrix factorization is shown in Fig.1. 

\begin{figure}[!ht]
	\centering
	\includegraphics[width=2.8in]{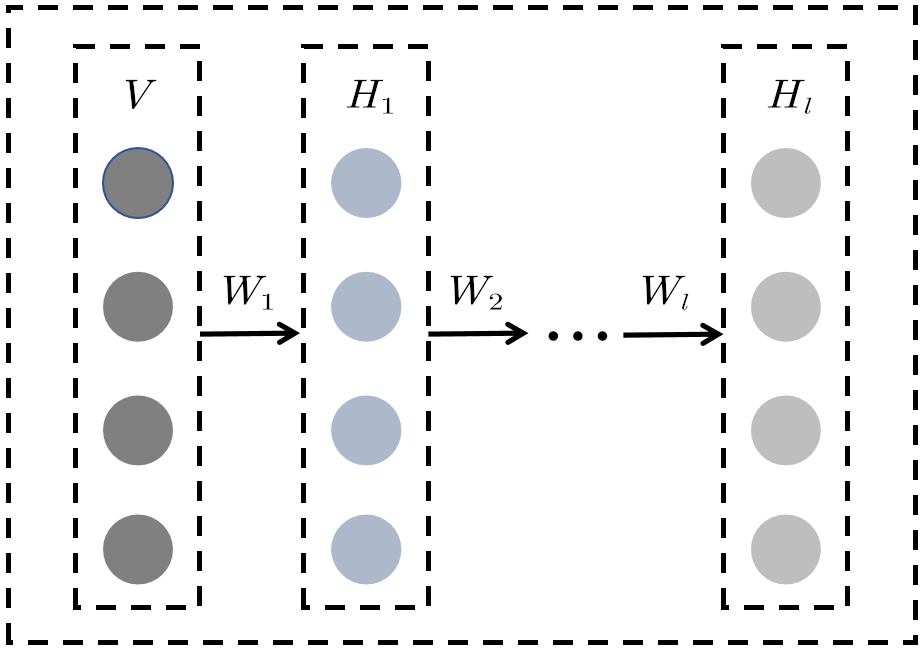}
	\caption{The structure of Deep NMF based on encoding matrix factorization.}
	\label{f1}
\end{figure}

The loss functions are shown in Eq.(1) and Eq.(2).
\begin{align}
	F &= \left\| V - W_1W_2\cdots W_lH_l \right\| ^2 \nonumber \\
	&= \sum_{i=1}^n{\sum_{j=1}^m{\left\| V_{ij} - \left( W_1W_2\cdots W_lH_l \right) _{ij} \right\| ^2}}
\end{align}
\begin{align}
	H_{k-1}=W_kH_k
\end{align}
where $1<k\leq l$.

The update rules for $W_{1}$ and $H_{1}$ are: 
\begin{align}
	W_1\gets W_1\odot \frac{VH_{1}^{T}}{VH_1H_{1}^{T}}
\end{align}
\begin{align}
	H_1\gets H_1\odot \frac{W_{1}^{T}V}{W_{1}^{T}W_1H_1}
\end{align}

The update rules for $W_{k}$ and $H_{k}$ are: 
\begin{align}
	W_k\gets W_k\odot \frac{H_{k-1}H_{k}^T}{H_{k-1}H_kH_k^T}
\end{align}
\begin{align}
	H_k\gets H_k\odot \frac{W_{k}^TH_{k-1}}{W_k^TW_kH_k}
\end{align}
Deep non-negative matrix factorization based on encoding matrix factorization is a method with broad application prospects. It performs well in feature extraction and data compression. However, during the process of layer-by-layer factorization, there are irreducible errors between layers, which may lead to a decrease in recognition rate with the increase of layers. Moreover, from the mathematical nature of NMF, non-negative matrix factorization is a non-convex optimization problem. The first-layer decomposition inherently yields a local optimum. Further decomposing the encoding matrix is equivalent to finding another local optimum to fit the previous local optimum. This cannot guarantee that the recognition rate will increase stably with the increase in the number of layers. In addition, the encoding matrix stores the weights of the basis images. Factorizing the encoding matrix is equivalent to factorizing the weights into the mixing matrix and encoding matrix of the next layer, where the mixing matrix stores the basis images. The concept of basis images in the coding matrix does not conform to human visual intuition, so the non-negative matrix factorization algorithm based on encoding matrix factorization has weak interpretability.

\subsection{Mixing maxtix-based Deep NMF}

To address the weak interpretability of deep NMF based on encoding matrix factorization, recent studies have adopted factorizing the mixing matrix instead of the encoding matrix. This method factorizes the mixing matrix $W_{k-1}$ from the previous layer into the mixing matrix $W_k$ and encoding matrix $H_k$ for the next layer. Consequently, all columns of the next layer's mixing matrix can be regarded as deeper features of the SAR image dataset. This approach significantly enhances the algorithm's interpretability without affecting its performance. The structure of NMF based on encoding matrix decomposition is shown in Fig.2. 

\begin{figure}[!ht]
	\centering
	\includegraphics[width=2.8in]{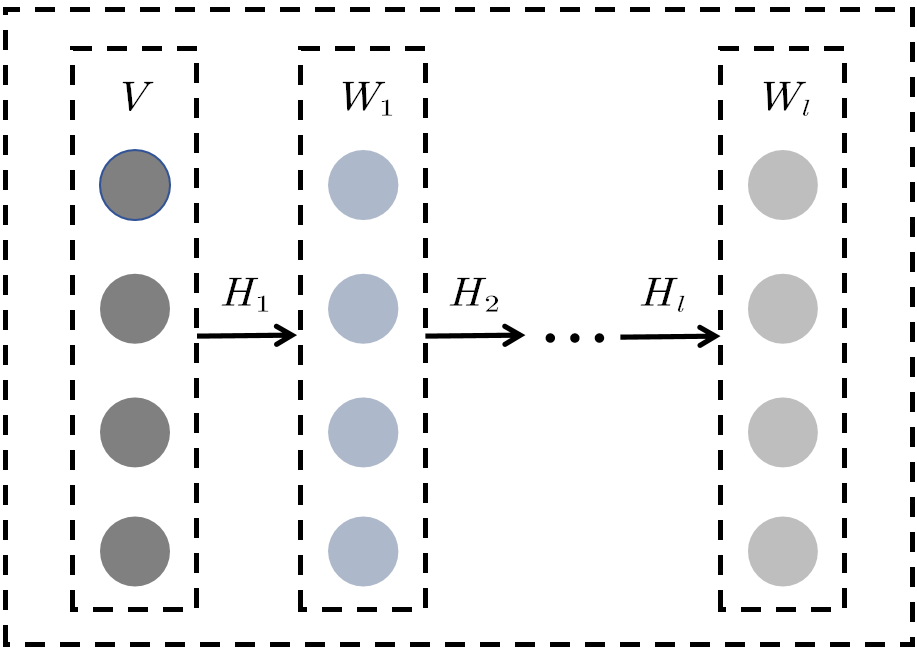}
	\caption{The structure of Deep NMF based on mixing matrix factorization.}
	\label{f2}
\end{figure}

The loss function for deep non-negative matrix factorization based on mixture matrix factorization is as follows:
\begin{align}
	F &= \left\| V - W_lH_lH_{l-1}\cdots H_1 \right\| ^2 \nonumber \\
	&= \sum_{i=1}^n{\sum_{j=1}^m{\left\| V_{ij} - \left( W_lH_lH_{l-1}\cdots H_1 \right) _{ij} \right\| ^2}}
\end{align}
\begin{align}
	W_{k-1}=W_kH_k
\end{align}
where $1<k\leq l$.

The update rules for $W_{1}$ and $H_{1}$ are same as deep NMF based on encoding matrix factorization.

The update rules for $W_{k}$ and $H_{k}$ are: 
\begin{align}
	W_k\gets W_k\odot \frac{W_{k-1}H_{k}^{T}}{W_{k-1}H_kH_{k}^{T}}
\end{align}
\begin{align}
	H_k\gets H_k\odot \frac{W_{k}^{T}W_{k-1}}{W_{k}^{T}W_kH_k}
\end{align}

Since the mixing matrix stores the basis images of the dataset, and these basis images are image features of the original images, deep non-negative matrix factorization (NMF) based on mixed decoding addresses to some extent the issue of weak interpretability in traditional methods. By decomposing the mixing matrix layer by layer into the mixing matrix and the encoding matrix of the next layer, the mixing matrix of the next layer can better represent the deeper-level features of the data. However, this method still has some limitations. As a layer-by-layer decomposition algorithm, it cannot solve the problems of cumulative errors and local optimality brought about by layer-by-layer decomposition. Therefore, although deep NMF based on mixing matrix decomposition has improved in terms of interpretability, further research is still needed to overcome its limitations.

Most deep non-negative matrix factorization algorithms are based on the above two layer-by-layer factorization architectures. However, layer-by-layer factorization causes these algorithms to have common limitations, including error accumulation, the unidirectional information transfer of matrices, and excessive dependence on the quality of the previous layer. To overcome these problems, this paper innovatively proposes a non-negative matrix factorization algorithm that avoids layer-by-layer factorization, as follows.

\section{Generalized Deep Non-negative Matrix Factorization }
This paper proposes a deep non-negative matrix factorization method based on a global perspective. During parameter updating, this method fully considers the relationships between the parameters being updated and the other parameters. The update of each matrix is determined collectively by all the other parameters. In the event of suboptimal parameter estimation, the global update mechanism ensures that the remaining parameters are adjusted cooperatively with the objective of minimizing the loss function, thereby reducing the negative impact caused by such local anomalies. Fig.3 shows the structure of G-DNMF.
\begin{figure}[!ht]
	\centering
	\includegraphics[width=2.8in]{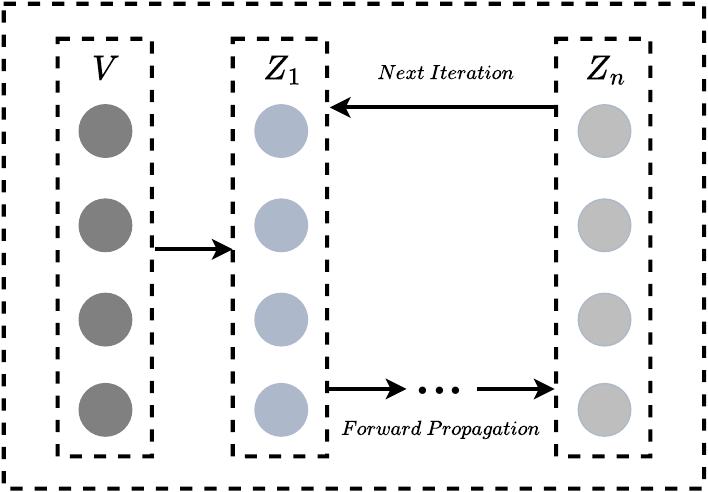}
	\caption{The structure of Generalized Deep Non-negative Matrix Factorization.}
	\label{f3}
\end{figure}

The features extracted by G-DNMF exhibit a distinct hierarchical structure. Shallow-level features are often more specific, with clear geometric and contour features that reflect the original images. In contrast, deep-level features are more abstract and often represent the intrinsic characteristics that are difficult to discover in the original dataset. Compared with the more concrete shallow-level features, these deep-level features better reflect the essential properties of the original dataset and thus possess stronger discriminative power. When used for classification with a classifier, deep-level features tend to perform better than shallow-level features. The proposed method consists of two steps. The first step involves performing a non-negative initialization of the matrices, and subsequently employing NMF to iterate on them. The second step employs the iterative formulas of G-DNMF to update each parameter until the convergence condition is met.

\subsection{The initialization of G-DNMF}
Without parameter initialization, G-DNMF is highly prone to vanishing gradients, as it is a highly complex non-convex optimization problem whose loss landscape contains numerous local optima. To avoid this problem, generalized deep non-negative matrix factorization must first be properly initialized, and the strategy adopted here employs non-negative matrix factorization based on mixed matrix factorization iterated only a few times. This choice is motivated by the fact that the mixed-matrix-factorization NMF framework offers excellent interpretability while ensuring the effectiveness of subsequent feature extraction, The expression is as follows.
\begin{align}
	 V  \approx  W_l  H_lH_{l-1}\cdots H_1
\end{align}
\begin{align}
	 V \approx  W_1  H_1
\end{align}
\begin{align}
	 W_{k-1}  \approx  W_k  H_k
\end{align}
where $1<k\leq l$.

Then, let $Z_{1}= W_l$, $Z_{2}=H_l$, $Z_{3}=H_{l-1}$, …, $Z_{l+1}=H_{1}$. Based on this, the next section will present the loss function for Generalized Deep Non-negative Matrix Factorization.
\subsection{The loss Function of G-DNMF}
The loss function of G-DNMF is given by Eq.(14).
\begin{align}
	F &=\left\| g\left( V \right) -Z_1Z_2\cdots Z_lZ_{l+1} \right\| ^2 \notag \\
	&=\sum_{i=1}^n{\sum_{j=1}^m{\left\| g\left( V \right) _{ij}-\left( Z_1Z_2\cdots Z_{l+1} \right) _{ij} \right\| ^2}}
\end{align}
where $g$ is the activation function.

The iterative formulas for the parameters in the loss function can be derived from the KKT conditions, and these formulas along with their derivations will be detailed in the subsequent subsection. In the above functions, the iterative formula for $Z_1$ is consistent with the update rule for the mixing matrix in NMF, while the iterative formula for $Z_{l+1}$ aligns with the update rule for the encoding matrix in NMF. Consequently, $Z_1$ acts as the mixture matrix in the algorithm, and $Z_{l+1}$ serves as the coding matrix. The iterative formulas for $Z_2, Z_3, \ldots, Z_l$ are more intricate. Both the encoding matrix and mixing matrix update rules in NMF can be seen as special cases of these parameter update formulas. Hence, this represents a more generalized iterative formula. Moreover, by abandoning layer-wise factorization, each parameter update aims to minimize the entire loss function. This means that the parameter values upon convergence are the final ones, eliminating the need for additional fine-tuning.

\subsection{The Update Rules of G-DNMF}
The loss function minimized by G-DNMF is shown in Eq. (13), and this mathematical optimization problem can be formulated as follows:
\begin{align}
	\min F &= \left\| g\left( V \right) - Z_1Z_2\cdots Z_lZ_{l+1} \right\| ^2 \notag \\
	&\text{s.t. } Z_i \geq 0 \quad \text{for } i = 1, 2, \cdots, l+1
\end{align}
To solve this optimization problem, this paper introduce the KKT (Karush-Kuhn-Tucker) conditions. First, this paper construct the Lagrangian function:
\begin{align}
	&L\left( Z_1, Z_2, \cdots , Z_{l+1}, \lambda _1, \lambda _2, \cdots , \lambda _{l+1} \right) \notag \\
	&=\left\| g(V)-Z_1Z_2\cdots Z_{l+1} \right\| ^2+\sum_{i=1}^{l+1}{\lambda _iZ_i}
\end{align}
Let
\begin{align}
	\alpha &= \prod_{k=1}^{i-1}{Z_i} \quad \beta = \prod_{k=i+1}^{l+2}{Z_i}
\end{align}
Next, take the partial derivative of \( L \) with respect to each \( Z_i \) and set it to zero:
\begin{align}
	\frac{\partial L}{\partial Z_i}=-2\alpha ^Tg(V)\beta ^T+2\alpha ^T\alpha Z_i\beta \beta ^T+\lambda _i=O 
\end{align}
From the complementary slackness condition, we have:
\begin{align}
	\lambda _i\odot Z_i=O
\end{align}
leading to:
\begin{align}
	\alpha ^Tg(V)\beta ^T\odot Z_i=\alpha ^T\alpha Z_i\beta \beta ^T\odot Z_i
\end{align}
From this, the iterative formula for G-DNMF can be derived as:
\begin{align}
	Z_i\gets Z_i\odot \frac{\alpha ^Tg(V)\beta ^T}{\alpha ^T\alpha Z_i\beta \beta ^T}
\end{align}
A detailed proof of the convergence of the G-DNMF iterative formula will be provided in the appendix.

The update rules of G-DNMF are highly flexible in form. As illustrated in Fig. 4, the formal consistency between the shaded regions in subfigures 1 and 2 indicates that the update formula for the encoding matrix can be considered a special case of the G-DNMF update formula. Similarly, the structural equivalence between the shaded regions in subfigures 3 and 4 demonstrates that the update formula for the mixing matrix can also be viewed as a special case of the G-DNMF update. Therefore, employing the G-DNMF update formula provides enhanced flexibility in parameter updating.

\begin{figure}[!ht]
	\centering
	\includegraphics[width=3.5in]{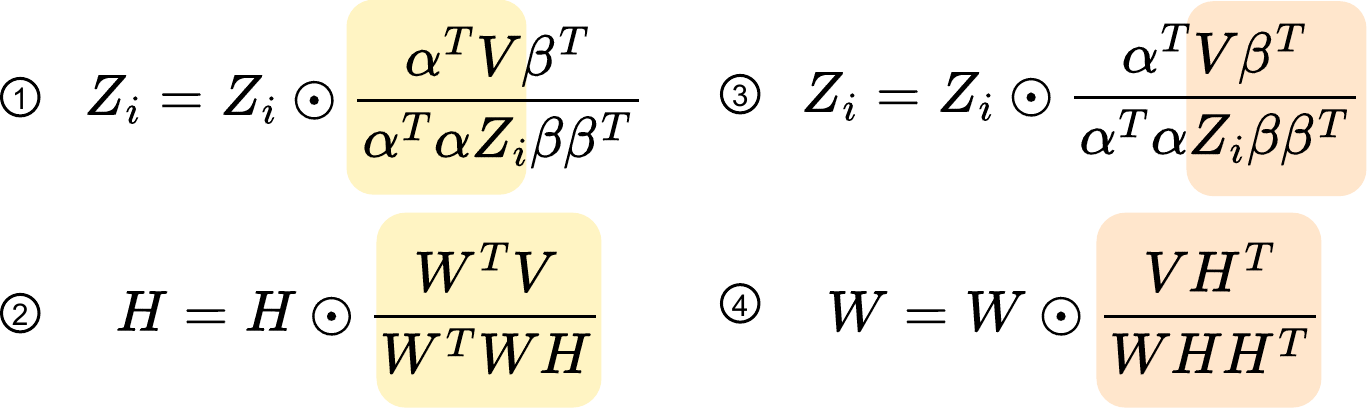}
	\caption{The relationship between the G-DNMF update rules and the update rules for the encoding matrix and the mixing matrix. (For better presentation, $g(V)$ is treated as $V$ here.)}
	\label{f3}
\end{figure}

\subsection{The Implementation Details of G-DNMF}
Similar to many deep non-negative matrix factorization methods, G-DNMF also has an initialization step. The initialization method used in this paper is nonlinear deep non-negative matrix factorization based on mixing matrix factorization, as shown in Eq. (11), where the nonlinear activation function $g$ is chosen as the $tanh$ function. This method can extract nonlinear higher-level features of SAR image datasets. However, we do not iterate until convergence but only perform 30 iterations. In fact, experiments show that 17 iterations during the initialization phase are sufficient to avoid gradient vanishing in G-DNMF.

After completing the initialization step, update the parameters starting from \(Z_1\) and proceeding sequentially backward. Repeat this process until the convergence condition is met. Experimental results show that the mixing matrices of each layer in the model exhibit discriminative effects. A comparison of the discriminative abilities of each layer will be provided in a subsequent part of the article. The pseudocode for this model process is given below.

\begin{algorithm}
	\caption{Generalized Deep Non-negative Matrix Factorization (G-DNMF)}
	\label{alg: 1}
	\begin{algorithmic}
		\REQUIRE {The raw data matrix $\mathbf{V}$, List of he number of features per layer, Number of layers in the model $\mathbf{n}$, $\mathbf{layers}$, The identity matrix $\mathbf{Z_0}$ and $\mathbf{Z_{n+1}}$, Maximum number of iterations $\mathbf{max}$}
		\ENSURE {The mixing matrices's cell $\mathbf{W}$, The encoding matrices's cell $\mathbf{H}$, The matrices's cell $\mathbf{Z}$}
		\FOR{i=1,...,$\mathbf{n}$}
		\IF{i=1}
		\STATE {$\mathbf{W}_1$, $\mathbf{H}_1$ $\gets$ $NMF$($\mathbf{V}$, $\mathbf{layers}$ (i))}
		\ELSE 
		\STATE {$\mathbf{W}_1$, $\mathbf{H}_1$ $\gets$ $NMF$ ($\mathbf{W}_{i-1}$, $\mathbf{layers}$ (i))}
		\ENDIF
		\ENDFOR
		\FOR{i=1,...,$\mathbf{n}+1$}
		\IF{i=1}
		\STATE {$\mathbf{Z}_i$=$\mathbf{W}_n$ }
		\ELSE 
		\STATE {$\mathbf{Z}_i$=$\mathbf{H}_{n-i+2}$}
		\ENDIF
		\ENDFOR
		\FOR{i=1,...,$\mathbf{max}$}
		\FOR{i=1,...,$\mathbf{n}+1$}
		\STATE{$\alpha$ = $\displaystyle\prod_{k=1}^{i-1}\mathbf{Z}_i \quad$}
		\STATE{$\beta$ = $\displaystyle\prod_{k=i+1}^{n+2}\mathbf{Z}_i$}
		\STATE{	$\mathbf{Z}_i\gets Z_i\odot \displaystyle \frac{\alpha ^Tg(V)\beta ^T}{\alpha ^T\alpha Z_i\beta \beta ^T}$}
		\ENDFOR
		\ENDFOR
	\end{algorithmic}
\end{algorithm} 

In the number of features per layer, this paper adopts a scheme where the number of features increases with the number of layers. The feature dimensions for the first through fourth layers are set to 32, 64, 128, and 256, respectively. The core idea is that the model first extracts more abstract features from the SAR image dataset, and then, as the number of layers increases, gradually extracts more concrete features. Concrete features can capture more detailed information, making the deep-level features more flexible in image reconstruction and better at preserving image detail information. Also, since the iterative formula of each layer involves the dataset V, unlike previous deep non-negative matrix factorization algorithms, the features extracted by this model at each layer are directly related to the dataset V, rather than indirectly related. Therefore, as the number of layers in the model increases, the basis images will not become distorted, thus enhancing the robustness and performance of the model.
\section{Experimental  Results}
In this study, we conducted experiments using both the MSTAR tank image dataset and the OpenSARShip dataset. The MSTAR dataset, widely used in SAR automatic target recognition, includes SAR images of various tank targets with added noise. For our experiments, we selected 64×64 pixel and 158×158 pixel raw data slices from ten distinct tank categories in the MSTAR dataset. The training samples were primarily composed of SAR targets with a pitch angle of 17 degrees, including a few with 30 degrees. For testing, we used SAR target samples with a pitch angle of 15 degrees to evaluate the model's generalization ability within the same domain. To further evaluate the model's performance and generalization capabilities across different datasets, the OpenSARShip dataset was additionally employed, from which we selected 64×64 pixel ship SAR images across four distinct categories. Fig. 5 shows the target images from both the MSTAR and OpenSARShip datasets.

\begin{figure}[!t]
	\centering
	%	% 第一行：10张图片（无标签）
	%	\includegraphics[width=0.65in]{t1}%
	%	\hfil
	%	\includegraphics[width=0.65in]{t2}%
	%	\hfil
	%	\includegraphics[width=0.65in]{t3}%
	%	\hfil
	%	\includegraphics[width=0.65in]{t4}%
	%	\hfil
	%	\includegraphics[width=0.65in]{t5}%
	%	\hfil
	%	\includegraphics[width=0.65in]{t6}%
	%	\hfil
	%	\includegraphics[width=0.65in]{t7}%
	%	\hfil
	%	\includegraphics[width=0.65in]{t8}%
	%	\hfil
	%	\includegraphics[width=0.65in]{t9}%
	%	\hfil
	%	\includegraphics[width=0.65in]{t10}%
	%	\\[5pt] % 换行并添加间距
	%	% 第二行：剩余10张图片
	\includegraphics[width=0.65in]{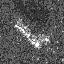}%
	\hfil
	\includegraphics[width=0.65in]{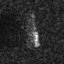}%
	\hfil
	\includegraphics[width=0.65in]{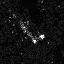}%
	\hfil
	\includegraphics[width=0.65in]{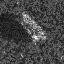}%
	\hfil
	\includegraphics[width=0.65in]{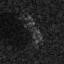}%
	\\[3pt]
	
	\includegraphics[width=0.65in]{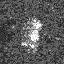}%
	\hfil
	\includegraphics[width=0.65in]{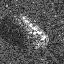}%
	\hfil
	\includegraphics[width=0.65in]{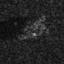}%
	\hfil
	\includegraphics[width=0.65in]{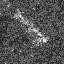}%
	\hfil
	\includegraphics[width=0.65in]{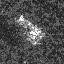}%
	\\[3pt]
	
	\includegraphics[width=0.65in]{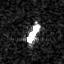}%
	\hfil
	\includegraphics[width=0.65in]{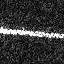}%
	\hfil
	\includegraphics[width=0.65in]{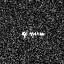}%
	\hfil
	\includegraphics[width=0.65in]{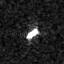}%
	\hfil
	
	\caption{The first two rows present Synthetic Aperture Radar images of tanks from the MSTAR dataset. These ten categories are arranged in order as 2S1, BMP2, BRDM, BTR60, BTR70, D7, T62, T72, ZIL131, and ZSU234. The bottom row displays Synthetic Aperture Radar images of ships from the OpenSARShip dataset, arranged from left to right as bulk carrier, container ship, general cargo, and tanker.}
	\label{best}
\end{figure}

The experimental design aims to comprehensively evaluate the performance and characteristics of the G-DNMF algorithm. Firstly, the interpretability of the algorithm is investigated, specifically focusing on the representation of features extracted by each layer when reconstructed in the original image, as well as the evolution of these features with increasing network depth. The features of each layer are visualized to demonstrate this interpretability. Secondly, the capability of G-DNMF to mitigate the performance degradation typically observed in traditional deep NMF algorithms as the number of layers increases is verified. By comparing performance indicators across different layers, the stability of G-DNMF in deep structures is evaluated. Thirdly, the feature dimension design of G-DNMF is experimentally explored to determine whether a dimension-incremental architecture or a Dimension-Reduction Architecture is more effective. Finally, the performance of G-DNMF is compared with that of other methods under varying sample data scales and across different datasets to assess its applicability and advantages under diverse data conditions. This is the first demonstration of systematic experiments on DNMF across multiple aspects including interpretability, dimension selection, layer configuration, and generalization performance, which is fundamentally underpinned by the theoretical guarantees of the G-DNMF derived in this paper.

\subsection{The Interpretability of Proposed Approach}
Non-negative Matrix Factorization (NMF) and its derivative algorithms possess an innate advantage in interpretability, stemming from the non-negativity constraint and the intuitive nature of their factorization results. Specifically, NMF decomposes the original data matrix $V$ into two low-rank non-negative matrices $W$ and $H$, such that $V \approx WH$. In tasks such as Synthetic Aperture Radar (SAR) image target recognition, where data is non-negative by nature, this decomposition preserves the physical meaning of the original data. The mixing matrix $W$ can be interpreted as foundational feature patterns, while the matrix $H$ represents the weight distribution of each image sample over these features. Since both matrices contain only non-negative elements, each data point is represented as a purely additive linear combination of basis vectors, strictly aligning with the human intuition that images are composed of combinations of distinct features.

Furthermore, NMF naturally tends to produce sparse factorization results, which significantly enhances its interpretability. Sparsity implies that the complexity of each feature constituting the image remains low, which is consistent with the human cognitive process of moving from simple to complex structures. In summary, the combination of non-negativity and sparsity enables NMF to reveal the intrinsic structure of data in a highly comprehensible manner. Building upon this solid foundation, Generalized Deep Non-negative Matrix Factorization (G-DNMF) extends the traditional framework by factorizing the dataset into several sequential matrices rather than just two, adapting the inherent transparency of NMF for multi-layered feature representation.

To demonstrate the interpretability of G-DNMF, an in-depth analysis of the features extracted by each layer is essential. As a deep feature extraction method, the interpretability of G-DNMF is reflected in its capacity to intuitively manifest the features learned by each layer and track their evolution as the algorithm depth increases. Specifically, the $i$-th layer features are derived through the sequential product of the factorized matrices, denoted as $Z_{1}Z_{2}\dots Z_{i}$. This multiplicative procedure enables a hierarchical visualization of the feature extraction process, demonstrating how the model progressively transitions from concrete geometric structures to abstract intrinsic properties. The following sections provide a detailed discussion on the interpretability of G-DNMF.

Firstly, starting with the analysis of the first layer, the features extracted by the first layer are the most vivid, and they may correspond to the overall structure or pattern in the image. These features are the basic units that constitute more complex patterns. By visualizing the feature mapping of the first layer, we can see how they capture the basic structure of the image. For example, in SAR images, the first layer may extract the overall contour of the target or its main geometric shape. These features provide the foundation for feature extraction in subsequent layers. The first row of Fig. 6 shows the corresponding features in the first layer.

As the number of layers increases, the features extracted by the subsequent layers of the model become increasingly abstract. The combination of shallow-layer features can be regarded as specific shapes or contours in SAR images. Deep-layer features may begin to capture more abstract characteristics of the target. Notably, although these deep-level features appear visually abstract, they still possess the capability to reconstruct the original images. Experimental results show that these abstract features often have a stronger discriminative effect. By comparing the features of each layer, the evolution of features from concrete to abstract can be observed. This evolution not only reflects the deepening of the model's understanding of the data but also shows the hierarchical nature of G-DNMF in feature extraction. The following presents the feature images extracted by G-DNMF from the second to the fourth layer.
%\begin{figure}[!h]
%	\centering
%	% 第一行：10张图片（无标签）
%	\includegraphics[width=0.233\linewidth]{a1}%
%	\hfill
%	\includegraphics[width=0.233\linewidth]{a2}%
%	\hfill
%	\includegraphics[width=0.233\linewidth]{a3}%
%	\hfill
%	\includegraphics[width=0.233\linewidth]{a4}%
%	\\[3pt] % 换行并添加间距
%	% 第二行：剩余10张图片
%	\caption{The features of the first layer in G-DNMF}
%	\label{layer1}
%\end{figure}
\begin{figure}[!htbp]
	\centering
	% 第一行：第1层特征
	\includegraphics[width=0.233\linewidth]{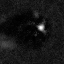}%
	\hfill
	\includegraphics[width=0.233\linewidth]{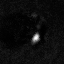}%
	\hfill
	\includegraphics[width=0.233\linewidth]{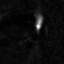}%
	\hfill
	\includegraphics[width=0.233\linewidth]{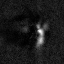}%
	\\[5pt] 
	
	% 第二行：第2层特征
	\includegraphics[width=0.233\linewidth]{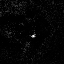}%
	\hfill
	\includegraphics[width=0.233\linewidth]{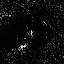}%
	\hfill
	\includegraphics[width=0.233\linewidth]{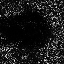}%
	\hfill
	\includegraphics[width=0.233\linewidth]{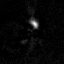}%
	\\[5pt] 
	
	% 第三行：第3层特征
	\includegraphics[width=0.233\linewidth]{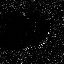}%
	\hfill
	\includegraphics[width=0.233\linewidth]{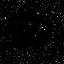}%
	\hfill
	\includegraphics[width=0.233\linewidth]{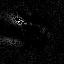}%
	\hfill
	\includegraphics[width=0.233\linewidth]{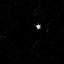}%
	\\[5pt] 
	
	% 第四行：第4层特征
	\includegraphics[width=0.233\linewidth]{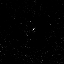}%
	\hfill
	\includegraphics[width=0.233\linewidth]{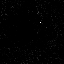}%
	\hfill
	\includegraphics[width=0.233\linewidth]{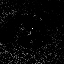}%
	\hfill
	\includegraphics[width=0.233\linewidth]{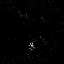}%
	
	\caption{The features extracted by G-DNMF from the first to the fourth layer are displayed from the top row to the bottom row, respectively. By comparing the features of different layers, it can be observed that as the depth of the model increases, the features evolve from being concrete to abstract.}
	\label{fig:combined_features}
\end{figure}

%\begin{figure}[!h] % 使用 H 固定位置（需要 float 宏包）
%	\centering
%	% 第一行：4张子图 (第2层特征)
%	\includegraphics[width=0.233\linewidth]{b1}%
%	\hfill
%	\includegraphics[width=0.233\linewidth]{b2}%
%	\hfill
%	\includegraphics[width=0.233\linewidth]{b3}%
%	\hfill
%	\includegraphics[width=0.233\linewidth]{b4}%
%	\\[5pt] % 第一行结束，换行并增加 5pt 间距
%	
%	% 第二行：4张子图 (第3层特征)
%	\includegraphics[width=0.233\linewidth]{b5}%
%	\hfill
%	\includegraphics[width=0.233\linewidth]{b6}%
%	\hfill
%	\includegraphics[width=0.233\linewidth]{b7}%
%	\hfill
%	\includegraphics[width=0.233\linewidth]{b8}%
%	\\[5pt] % 第二行结束，换行并增加 5pt 间距
%	
%	% 第三行：4张子图 (第4层特征)
%	\includegraphics[width=0.233\linewidth]{b9}%
%	\hfill
%	\includegraphics[width=0.233\linewidth]{b10}%
%	\hfill
%	\includegraphics[width=0.233\linewidth]{b11}%
%	\hfill
%	\includegraphics[width=0.233\linewidth]{b12}%
%	
%	\caption{The first to third rows correspond to the features of the second to fourth layers in G-DNMF, respectively. By comparing the features of different layers in G-DNMF, it can be observed that as the depth of the model increases, the features evolve from being more concrete to more abstract.}
%	\label{fig:b1-b12}
%\end{figure}

Since all elements in the matrix are positive values, the proposed method can satisfy the additivity principle during the original image reconstruction process. This characteristic is consistent with the objective laws of human cognition and demonstrates the interpretability of the algorithm.

G-DNMF employs a global optimization strategy, effectively avoiding inter-layer errors that typically arise in layer-wise factorization. Unlike traditional deep non-negative matrix factorization methods, G-DNMF ensures that the feature reconstructions of each layer remain visually consistent. This consistency is achieved because there is no distortion caused by the propagation of errors through the layers. As a result, the discriminative effects of each layer's features are determined purely by the intrinsic properties of the features themselves, without any interference from noise. Detailed experimental data and in-depth analysis of the discriminative effects of each layer will be provided in the subsequent sections of this paper. Below, is the feature reconstruction maps generated by G-DNMF.

\begin{figure}[H] % 使用 H 浮动位置
	\centering
	% 第一行：4张子图
	\includegraphics[width=0.233\linewidth]{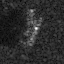}%
	\hfill
	\includegraphics[width=0.233\linewidth]{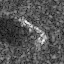}%
	\hfill
	\includegraphics[width=0.233\linewidth]{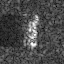}%
	\hfill
	\includegraphics[width=0.233\linewidth]{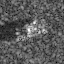}%
	\\[5pt] % 换行
	% 第二行：4张子图
	\includegraphics[width=0.233\linewidth]{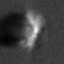}%
	\hfill
	\includegraphics[width=0.233\linewidth]{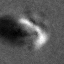}%
	\hfill
	\includegraphics[width=0.233\linewidth]{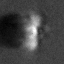}%
	\hfill
	\includegraphics[width=0.233\linewidth]{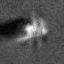}%
	\caption{The first row presents the original images, while the second row displays the images reconstructed using G-DNMF. It can be observed that the reconstructed images exhibit a close correspondence in shape and structure to the original images.} % 图片标题
	\label{fig:c1-c8} % 图片标签
\end{figure}

\subsection{The Architectural Design of Proposed Approach}
In deep feature extraction models, the selection of the number of features per layer is crucial as it directly impacts the model's representation and generalization abilities. Regarding the architectural design of deep non-negative matrix factorization, there are two main design philosophies: one where the feature dimension decreases as the number of layers increases, and another where the feature dimension increases with the number of layers. Below is the analysis of these two approaches presented in this paper.

Both the dimension-reduction and dimension-incremental architectures present reasonable strategies for deep feature extraction. The dimension-incremental approach provides a richer representational space to capture complex high-order patterns, though it inherently incurs higher computational costs and an increased risk of overfitting. Conversely, the dimension-reduction approach effectively mitigates overfitting and lowers memory demands, but it may restrict the model's capacity to preserve fine details in complex datasets.
%In summary, the architecture with decreasing feature dimensions is suitable for scenarios requiring efficient computation and low data complexity. It reduces the risk of overfitting and enhances the model's generalization ability. However, it may lead to information loss and limited representation for highly complex datasets. Conversely, the architecture with increasing feature dimensions is ideal for complex datasets due to its superior representation and deep feature extraction capabilities but is not recommended when computational efficiency and low data complexity are priorities.

To determine which architecture is more suitable for G-DNMF, this paper conducts a series of experiments to evaluate the performance of the G-DNMF algorithm under different architectural designs, with the corresponding results illustrated in Fig. 8. Specifically, when there are 400 samples, the average recognition rate of the dimension-reduction architecture is 86.244\%, while the average recognition rate of the dimension-increment architecture is 84.140\%. The recognition rate of the dimension-reduction architecture is 2.104\% higher than that of the dimension-increment architecture. When the number of samples increases to 2747, the dimension-reduction architecture achieves an average recognition rate of 94.699\% with a standard deviation of 0.314. In contrast, the dimension-increment architecture achieves a higher average recognition rate of 95.111\% along with a lower standard deviation of 0.175. At this point, the dimension-increment architecture not only outperforms the dimension-reduction architecture by 0.412\% in recognition rate but also exhibits a standard deviation that is 0.139 lower, indicating superior stability on larger datasets.
\begin{figure}[!ht]
	\centering
	\includegraphics[width=8cm]{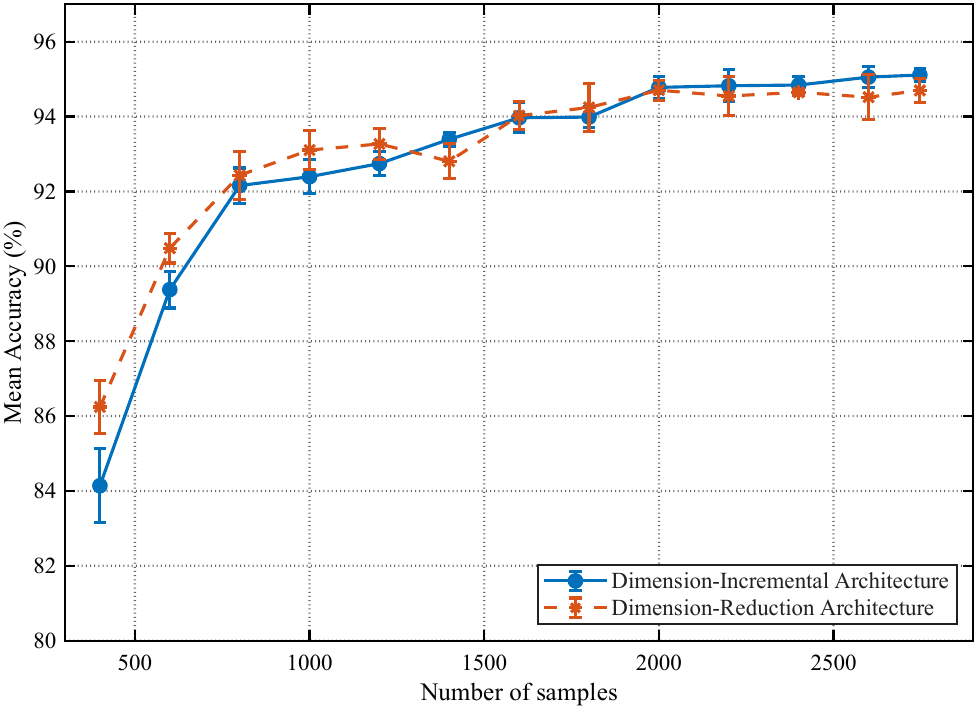}
	\caption{The variation of recognition rate with the increase of samples for different architectures. In the Dimension-Incremental Architecture, the feature dimensions of each layer from the first to the fourth layer are 32, 64, 128, and 256 respectively. In contrast, for the Dimension-Reduction Architecture, the feature dimensions of each layer from the first to the fourth layer are 256, 128, 64, and 32 respectively.}
	\label{f2}
\end{figure}

The experimental results indicate that with 400 samples, the dimension-reduction architecture performs better than the dimension-increment architecture because it is more adept at handling data with lower complexity. However, with 2747 samples, the dimension-increment architecture demonstrates superior performance on the SAR image dataset compared to the dimension-reduction architecture by achieving both a higher recognition rate and a smaller variance. This dual advantage can be attributed to the fact that the dimension-incremental architecture exhibits greater adaptability and robust stability when handling larger and more complex training datasets.

\subsection{The Stability of Proposed Approach}
To verify the stability of G-DNMF, it is essential to first understand what stability implies within the context of deep non-negative matrix factorization models: the model's performance should not degrade as the number of layers increases. This presents a significant challenge in many deep NMF architectures, as increasing the network depth can lead to issues such as feature distortion, error accumulation, local optima, and overfitting, all of which negatively impact performance. However, the design of G-DNMF incorporates key mechanisms to ensure performance remains stable and even improves with additional layers. Primarily, G-DNMF completely avoids error accumulation by employing a global optimization strategy rather than traditional layer-wise factorization. This approach eliminates the propagation of errors between layers, which is a common issue in layer-wise methods that severely degrades performance. By deriving parameter updates from a global perspective, G-DNMF ensures no inter-layer error accumulation occurs, allowing the converged results to directly serve as the final output. This global optimization mechanism is pivotal for maintaining stability as the network grows deeper.

Furthermore, G-DNMF ensures stability in feature extraction by employing a collaborative update mechanism that strictly anchors all extracted features to the original data across the entire process. In traditional layer-wise methods, features often become increasingly disconnected from the original data as the network deepens, which can lead to feature distortion and local optima. G-DNMF overcomes this vulnerability through its unique parameter update rules, where every parameter update involves continuous collaboration with all other parameters. Because each layer's update is influenced by both preceding and subsequent layers, later layers can adjust globally even if earlier layers perform poorly, while simultaneously providing feedback to guide earlier layers in the next cycle. More importantly, the original dataset directly participates in the update formula for every parameter. This direct participation guarantees that the features of every layer remain firmly anchored to the original data's essential characteristics during the updates, completely preventing feature distortion as the depth increases. In summary, the ability of G-DNMF to extract deep-level features through collaborative global updates guarantees both high interpretability and robust stability, distinguishing it from traditional deep NMF methods.

In addition to theoretical explanations of G-DNMF's stability, experimental verification is also crucial. This paper designs a series of experiments to compare G-DNMF with traditional algorithms, demonstrating that G-DNMF maintains or even enhances performance as layers increase, outperforming other algorithms in this aspect. The experiments utilize SAR image datasets for feature extraction, and the extracted features are subsequently evaluated by a classifier, with classification accuracy serving as the evaluation metric.

In the experiments, we compare G-DNMF with several feature extraction algorithms, including Deep NMF based on mixing matrix factorization, Deep NMF based on encoding matrix factorization, Deep RNMF based on mixing matrix factorization, Deep RNMF based on encoding matrix factorization, and Deep IRNMF. The features extracted by all algorithms are evaluated using a unified classifier, and the recognition rate is calculated as the average of five independent experimental runs. The figure illustrates how these average recognition rates vary with increasing network depth.

\begin{figure}[!ht]
	\centering
	\includegraphics[width=8cm]{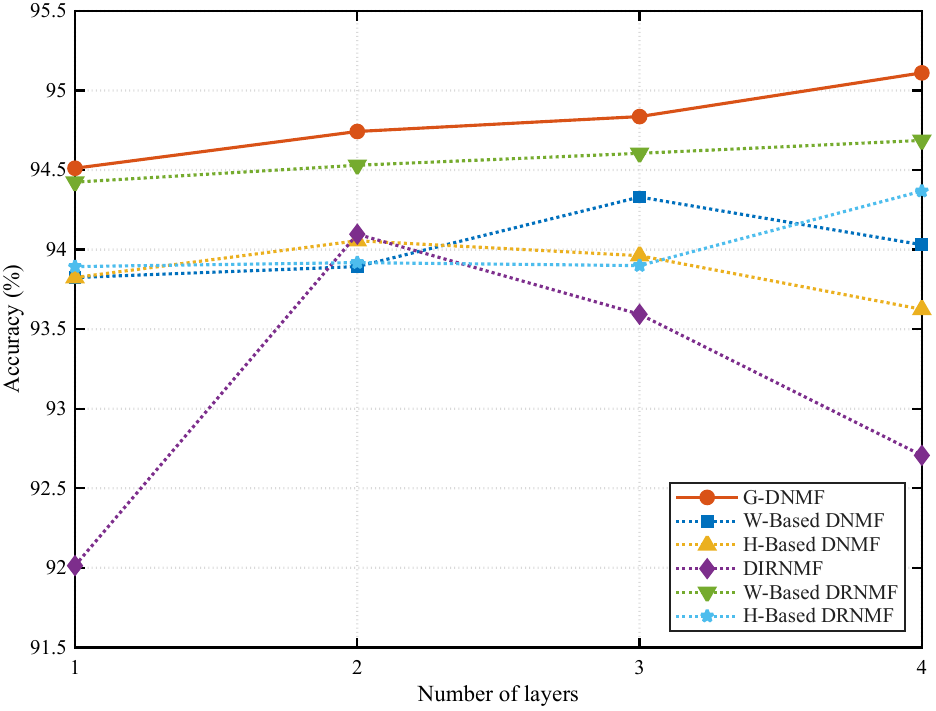}
	\caption{The accuracy of different NMF variant algorithms with the number of network layers variation.}
	\label{f2}
\end{figure}

As shown in the figure, when the number of network layers reaches 4, the recognition rate of G-DNMF achieves 95.111\%, which is higher than those of all other compared algorithms. Experimental results verify G-DNMF's stability as layers increase. Using a global optimization strategy and parameter update mechanism, G-DNMF avoids error accumulation from layer-wise factorization. This lets it maintain good performance in deep structures, giving it a clear edge over other deep NMF methods in image target recognition.

\subsection{The Recognition Performance of Proposed Approach}
To comprehensively evaluate the recognition performance and generalization ability of the proposed G-DNMF, a series of comparative experiments were conducted on multiple distinct datasets—namely, the $64 \times 64$ MSTAR, OpenSARShip, and $128 \times 128$ MSTAR datasets—under various sample sizes. The proposed model is benchmarked against several representative Deep NMF variants, including mixing-matrix-based and encoding-matrix-based Deep NMF (denoted as W-Based DNMF and H-Based DNMF), their robust counterparts (W-Based DRNMF and H-Based DRNMF), and Deep IRNMF (DIRNMF). For a fair comparison, the features extracted by all evaluated algorithms are fed into a unified classifier. To ensure statistical reliability, the average classification accuracy over five independent experimental runs is adopted as the primary performance metric.

\begin{table}[htbp]
	\centering
	\caption{Comparison of Recognition Rates on the $64 \times 64$ MSTAR Dataset at Different Sample Sizes: Mean (Std) (\%)}
	\label{tab:results_comparison_true_std_2decimals}
	\small 
	\begin{tabular}{l ccc}
		\toprule
		\multirow{2}{*}{\textbf{Algorithm}} & \multicolumn{3}{c}{\textbf{Sample Size}} \\
		\cmidrule(lr){2-4}
		& 2400 & 2600 & 2747 \\
		\midrule
		\textbf{G-DNMF} & \textbf{94.84 (0.24)} & \textbf{95.05 (0.28)} & \textbf{95.11 (0.17)} \\
		W-Based DRNMF & 94.37 (0.34) & 94.49 (0.59) & 94.74 (0.19) \\
		W-Based DNMF  & 93.97 (0.20) & 93.99 (0.56) & 94.03 (0.66) \\
		H-Based DRNMF & 93.81 (0.27) & 93.82 (0.11) & 93.92 (0.16) \\
		H-Based DNMF  & 93.62 (0.48) & 93.75 (0.49) & 94.06 (0.29) \\
		DIRNMF        & 93.49 (0.18) & 93.80 (0.24) & 94.10 (0.11) \\
		\bottomrule
	\end{tabular}
\end{table}

Table~\ref{tab:results_comparison_true_std_2decimals} presents the experimental results on the $64 \times 64$ MSTAR dataset. It can be observed that when the training sample size reaches 2747, the proposed G-DNMF achieves a peak recognition rate of 95.11\%, outperforming all baseline algorithms. Specifically, it yields performance margins of 0.37\%, 1.08\%, 1.01\%, 1.05\%, and 1.19\% over W-Based DRNMF, W-Based DNMF, DIRNMF, H-Based DNMF, and H-Based DRNMF, respectively. Additionally, the consistently small standard deviation underscores the superior stability and robustness of our approach.

\begin{table}[htbp]
	\centering
	\caption{Comparison of Recognition Rates on the OpenSARShip Dataset at Different Sample Sizes: Mean (Std) (\%)}
	\label{tab:results_opensarship}
	\small 
	\begin{tabular}{l ccc}
		\toprule
		\multirow{2}{*}{\textbf{Algorithm}} & \multicolumn{3}{c}{\textbf{Sample Size}} \\
		\cmidrule(lr){2-4}
		& 700 & 800 & 845 \\
		\midrule
		\textbf{G-DNMF} & \textbf{94.35 (0.22)} & \textbf{94.60 (0.29)} & \textbf{95.15 (0.29)} \\
		W-Based DRNMF & 94.10 (0.45) & 94.50 (0.73) & 94.90 (0.38) \\
		W-Based DNMF  & 93.70 (0.41) & 94.35 (0.65) & 94.35 (0.45) \\
		H-Based DRNMF & 93.45 (0.41) & 93.70 (0.48) & 94.25 (0.40) \\
		H-Based DNMF  & 93.80 (0.78) & 93.85 (0.55) & 93.95 (0.65) \\
		DIRNMF        & 91.80 (1.63) & 92.35 (0.82) & 93.75 (1.29) \\
		\bottomrule
	\end{tabular}
\end{table}

Similar superiority is evident on the OpenSARShip dataset, as detailed in Table~\ref{tab:results_opensarship}. Under the maximum evaluated sample size of 845, G-DNMF delivers the highest accuracy at 95.15\%. Compared to the existing NMF variants, our method exhibits performance improvements ranging from 0.25\% (against W-Based DRNMF) to 1.40\% (against DIRNMF). The narrow standard deviation further confirms the reliability of the proposed model in handling this dataset.

\begin{table}[htbp]
	\centering
	\caption{Comparison of Recognition Rates on the $158 \times 158$ MSTAR Dataset at Different Sample Sizes: Mean (Std) (\%)}
	\label{tab:results_mstar_128}
	\small 
	\begin{tabular}{l ccc}
		\toprule
		\multirow{2}{*}{\textbf{Algorithm}} & \multicolumn{3}{c}{\textbf{Sample Size}} \\
		\cmidrule(lr){2-4}
		& 2400 & 2600 & 2746 \\
		\midrule
		\textbf{G-DNMF} & \textbf{96.87 (0.65)} & \textbf{97.12 (0.31)} & \textbf{97.40 (0.29)} \\
		W-Based DRNMF & 96.55 (0.34) & 97.03 (0.23) & 97.17 (0.60) \\
		W-Based DNMF  & 96.65 (0.40) & 97.01 (0.52) & 97.03 (0.31) \\
		H-Based DRNMF & 96.57 (1.17) & 96.86 (0.62) & 96.95 (0.30) \\
		H-Based DNMF  & \textbf{96.87 (0.52)} & 96.89 (0.40) & 96.96 (0.39) \\
		DIRNMF        & 94.96 (0.69) & 95.60 (1.34) & 95.67 (0.35) \\
		\bottomrule
	\end{tabular}
\end{table}

To further evaluate the model's scalability on higher-resolution inputs, Table~\ref{tab:results_mstar_128} compares the recognition rates on the $128 \times 128$ MSTAR dataset. At a sample size of 2746, G-DNMF continues to dominate, reaching an impressive accuracy of 97.40\%. This translates to advantages of 0.23\%, 0.37\%, 0.44\%, 0.45\%, and 1.73\% over W-Based DRNMF, W-Based DNMF, H-Based DNMF, H-Based DRNMF, and DIRNMF, respectively. The persistently low variance highlights the model's resilience across different resolutions.

Overall, the experimental results across various datasets and different image resolutions indicate that the performance of G-DNMF steadily improves as the number of training samples increases, reflecting a highly favorable learning curve. The superior performance can be attributed to the proposed global optimization strategy and the architecture design that expands feature dimensions. These mechanisms effectively extract discriminative features while preserving structural stability, thereby ensuring strong generalization capabilities within deep architectures. Compared to other baseline algorithms, G-DNMF demonstrates enhanced adaptability and consistency when processing complex data with varying scales and across different domains, offering a highly effective and robust solution for deep feature representation and complex target recognition tasks.

%The experimental results indicate that the performance of G-DNMF steadily improves with the increase in the number of training samples, demonstrating a good learning curve and stability. Moreover, when the number of training samples exceeds 2000, G-DNMF consistently outperforms other non-negative matrix factorization (NMF) and its derivative models in terms of recognition performance. Its superior performance is attributed to the global optimization strategy and the architecture design of increasing feature dimensions, which can effectively extract discriminative features while maintaining stability and still exhibit good generalization ability in deep structures. Compared with other algorithms, G-DNMF shows higher stability and adaptability in handling complex datasets, providing an effective solution for SAR image target recognition tasks.

\section*{Acknowledgments}
This work was supported by the National Science and Technology Major Projects of China(2025ZD1008203), the National Natural Science Foundation of China(42572389), the Sichuan Natural Science Foundation(2026NSFSC0238)

{\appendix[The proof of the convergence of G-DNMF]
	
\subsection{Equivalent Expression}
The loss function and iterative formula of G-DNMF have been given in Eq.17 and Eq.21. In the following, the paper will provide the proof of convergence for the iterative formula of G-DNMF. The proposed method decomposes the original dataset \( \mathbf{V} \) into \( n \) matrices (\( n \geq 3 \)). The update formulas for the first and the \( n \)-th matrices are identical to the update formulas of NMF for the mixing matrix and the encoding matrix, respectively. Detailed proofs of convergence already exist, and thus will not be reiterated here. And for the \(i\)-th matrix (\(1 < i < n\)), its iterative formula multiplies the first \(i-1\) matrices together as a new matrix and the remaining \(n-i\) matrices as another new matrix. Therefore, the proof of convergence for the iterative formula when decomposing the original dataset into \(n\) matrices is essentially the same as the proof of convergence when decomposing it into 3 matrices. Hence, the following proof will be simplified to the case of decomposing into 3 matrices. Since the activation of the dataset has no effect on the convergence of the iterative formula, the loss function can be simplified as:
\begin{align}
	F=\frac{1}{2}\left\| V-AZB \right\| ^2
\end{align}
The loss function is multiplied by \(\frac{1}{2}\) for the convenience of subsequent differentiation. Here, \(V \in \mathbb{R}^{m \times n}\), \(A \in \mathbb{R}^{m \times p}\), \(Z \in \mathbb{R}^{p \times l}\), and \(B \in \mathbb{R}^{l \times n}\). Now, let's examine the convergence of the iterative formula for \(Z\).
 
\subsection{Proof of Convergence}
The loss function is constructed as follows:
\begin{align}
	F\left( Z_i \right) &= \frac{1}{2}\left\| V-AZB \right\|^2 \nonumber \\ 
	&= \frac{1}{2}\mathrm{tr}\left( \left( V^T-(AZB)^T \right) \left( V-AZB \right) \right)
	\label{eq:loss_func}
\end{align}
where $Z_i$ represents the $i$-th column of matrix $Z$.
The Taylor expansion of $F(Z_i)$ at $Z_i^t$ is given by:
\begin{align}
	F\left( Z_i \right) &= F\left( Z_{i}^{t} \right) + \left( Z_i-Z_{i}^{t} \right)^T \nabla F\left( Z_{i}^{t} \right) \nonumber \\
	&\quad + \frac{1}{2}\left( Z_i-Z_{i}^{t} \right)^T A^TA W_{ii} \left( Z_i-Z_{i}^{t} \right)
	\label{eq:taylor_expansion}
\end{align}
where $\nabla F\left( Z_{i}^{t} \right) = -A^TV B_i + A^TA Z_{i}^{t} W_{ii}$, and $W = BB^T$.
The auxiliary function is constructed as:
\begin{align}
	G\left( Z,Z_{i}^{t} \right) &= F\left( Z_{i}^{t} \right) + \left( Z-Z_{i}^{t} \right)^T \nabla F\left( Z_{i}^{t} \right) \nonumber \\
	&\quad + \frac{1}{2}\left( Z-Z_{i}^{t} \right)^T K\left( Z_{i}^{t} \right) \left( Z-Z_{i}^{t} \right)
	\label{eq:auxiliary_func}
\end{align}
where

\begin{equation}
	K\left( Z_{i}^{t} \right) = \left(
	\begin{array}{@{}c@{\hspace{-2pt}}c@{\hspace{-2pt}}c@{\hspace{-2pt}}c@{}}
		\tfrac{(A^T\!A Z_{i}^{t} W_{ii})_1}{(Z_{i}^{t})_1} & 0 & \cdots & 0 \\
		0 & \tfrac{(A^T\!A Z_{i}^{t} W_{ii})_2}{(Z_{i}^{t})_2} & \cdots & 0 \\
		\vdots & \vdots & \ddots & \vdots \\
		0 & 0 & \cdots & \tfrac{(A^T\!A Z_{i}^{t} W_{ii})_k}{(Z_{i}^{t})_k}
	\end{array}
	\right)
\end{equation}

Since the Hessian matrix of $G\left(Z,Z_{i}^{t}\right)$ is positive semi-definite, 
$G\left(Z_{i}^{t+1}, Z_{i}^{t}\right)$ attains its minimum when 
$\nabla G\left(Z_{i}^{t+1}, Z_{i}^{t}\right) = 0$. Thus, $G$ is monotonically decreasing.
If the inequality 
\begin{equation}
	F\left(Z_{i}^{t+1}\right) \leq G\left(Z_{i}^{t+1}, Z_{i}^{t}\right) \leq G\left(Z_{i}^{t}, Z_{i}^{t}\right) = F\left(Z_{i}^{t}\right)
\end{equation} 
holds, then $G$ has a lower bound 0. By the monotone convergence theorem, $G$ converges. 
Similarly, $F$ is also monotonically decreasing and bounded below by 0, hence $F$ converges.
The equalities $G\left(Z_{i}^{t}, Z_{i}^{t}\right) = F\left(Z_{i}^{t}\right)$ and 
$G\left(Z_{i}^{t+1}, Z_{i}^{t}\right) \leq G\left(Z_{i}^{t}, Z_{i}^{t}\right)$ are trivial. 
Thus, we only need to prove:
\begin{equation}
	F\left(Z_{i}^{t+1}\right) \leq G\left(Z_{i}^{t+1}, Z_{i}^{t}\right).
\end{equation}
Consider the difference:
\begin{multline}
	G\left(Z_{i}^{t+1},Z_{i}^{t}\right) - F\left(Z_{i}^{t+1}\right) \\
	= \frac{1}{2}\left(Z_{i}^{t+1}-Z_{i}^{t}\right)^T 
	\left(K\left(Z_{i}^{t}\right) - A^TAW_{ii}^T\right) 
	\left(Z_{i}^{t+1}-Z_{i}^{t}\right).
\end{multline}

If $\left(K\left(Z_{i}^{t}\right) - A^TAW_{ii}^T\right)$ is positive semi-definite, then
\begin{equation}
	G\left(Z_{i}^{t+1},Z_{i}^{t}\right) - F\left(Z_{i}^{t+1}\right) \geq 0
\end{equation}
holds. Let:
\begin{equation}
	M = D \left( K\left( Z_{i}^{t} \right) - A^TAW_{ii}^T \right) D
	\label{eq:M_compact}
\end{equation}
where $D = \mathrm{diag}(Z_i^t)$ is the diagonal matrix whose elements are $\left(Z_i^t\right)_1, \left(Z_i^t\right)_2, \dots, \left(Z_i^t\right)_k$.

For any vector $C$, if $C^TMC \geq 0$ holds, then $\left( K\left( Z_{i}^{t} \right) - A^TAW_{ii}^T \right)$ is positive semi-definite. 

\begin{align}
	C^TMC &= \sum_{a,b} C_a M_{ab} C_b \nonumber \\
	&= \sum_{a,b} C_a \left( Z_{i}^{t} \right)_a K\left( Z_{i}^{t} \right)_{ab} \left( Z_{i}^{t} \right)_b C_b \nonumber \\
	&\quad - C_a \left( Z_{i}^{t} \right)_a \left( A^TAW_{ii}^T \right)_{ab} \left( Z_{i}^{t} \right)_b C_b
\end{align}
Since $K\left( Z_{i}^{t} \right)_{ab} = 0$ when $a \neq b$, we have:
\begin{align}
	\sum_{a,b} & C_a \left( Z_{i}^{t} \right)_a K\left( Z_{i}^{t} \right)_{ab} \left( Z_{i}^{t} \right)_b C_b \nonumber \\
	&= \sum_a C_a \left( Z_{i}^{t} \right)_a \frac{\left( A^TA Z_{i}^{t} W_{ii}^T \right)_a}{\left( Z_{i}^{t} \right)_a} \left( Z_{i}^{t} \right)_a C_a \nonumber \\
	&= \sum_a C_a^2 \left( A^TA Z_{i}^{t} W_{ii}^T \right)_a \left( Z_{i}^{t} \right)_a \nonumber \\
	&= \sum_{a,b} \left( \frac{1}{2}C_a^2 + \frac{1}{2}C_b^2 \right) \left( A^TAW_{ii}^T \right)_{ab} \left( Z_{i}^{t} \right)_b \left( Z_{i}^{t} \right)_a
\end{align}

\begin{align}
	\sum_{a,b} & C_a \left( Z_{i}^{t} \right)_a \left( A^TAW_{ii}^T \right)_{ab} \left( Z_{i}^{t} \right)_b C_b \nonumber \\
	&= \sum_{a,b} C_a C_b \left( A^TAW_{ii}^T \right)_{ab} \left( Z_{i}^{t} \right)_b \left( Z_{i}^{t} \right)_a
\end{align}
Therefore:
\begin{align}
	C^TMC &= \sum_{a,b} \left( \frac{1}{2}C_a^2 + \frac{1}{2}C_b^2 - C_a C_b \right) \nonumber \\
	&\quad \times \left( A^TAW_{ii}^T \right)_{ab} \left( Z_{i}^{t} \right)_b \left( Z_{i}^{t} \right)_a \nonumber \\
	&= \frac{1}{2} \sum_{a,b} \left( C_a - C_b \right)^2 \left( A^TAW_{ii}^T \right)_{ab} \nonumber \\
	&\quad \times \left( Z_{i}^{t} \right)_b \left( Z_{i}^{t} \right)_a \geq 0
\end{align}

This proves $M$ is positive semi-definite, and consequently $\left( K\left( Z_{i}^{t} \right) - A^TAW_{ii}^T \right)$ is positive semi-definite. Thus, $F$ converges.

When $\nabla G\left( Z_{i}^{t+1}, Z_{i}^{t} \right) = 0$, we have:
\begin{equation}
	\nabla F\left( Z_{i}^{t} \right) + K\left( Z_{i}^{t} \right) \left( Z_{i}^{t+1} - Z_{i}^{t} \right) = 0
\end{equation}
\subsection{Convergence Analysis}
Thus, the update rule becomes:
\begin{equation}
	Z_{i}^{t+1} = Z_{i}^{t} - K\left( Z_{i}^{t} \right)^{-1} \nabla F\left( Z_{i}^{t} \right)
	\label{eq:update_rule}
\end{equation}

which yields the element-wise multiplicative update:
\begin{equation}
	Z_{i}^{t+1} = \left( \frac{A^TV B_i^T}{A^TA Z_{i}^{t} W_i^T} \right) \odot Z_{i}^{t}
	\label{eq:multiplicative_update}
\end{equation}

where $\odot$ denotes the Hadamard (element-wise) product. This update rule guarantees:
\begin{itemize}
	\item Column-wise convergence of $Z$ during iterations
	\item Global convergence of the entire matrix $Z$
	\item Non-negativity preservation (if $Z_i^t$ is initialized as non-negative)
\end{itemize}


\end{document}